\pdfoutput=1

\documentclass[11pt]{article}

\usepackage[final]{ACL2023}

\usepackage{times}
\usepackage{latexsym}

\usepackage[T1]{fontenc}

\usepackage[utf8]{inputenc}

\usepackage{microtype}

\usepackage{inconsolata}

\usepackage{twemojis}
\usepackage{xcolor}
\usepackage{soul}
\usepackage{tabularray} 
\usepackage{graphicx}
\usepackage{tcolorbox}

\usepackage{fdsymbol}
\usepackage{cleveref}
\usepackage{booktabs}
\usepackage{amsmath}
\usepackage{multirow}

\title{Mitigating Frequency Bias and Anisotropy in Language Model Pre-Training with Syntactic Smoothing}

\newcommand{\norm}[1]{\left\lVert#1\right\rVert}
\newcommand\cincludegraphics[2][]{\raisebox{-0.2\height}{\includegraphics[#1]{#2}}}

\author{
    {\bf Richard Diehl Martinez} \texttwemoji{orange}~~~~
    {\bf Z\'{e}bulon Goriely} \texttwemoji{orange} ~~~~ \\
    {\bf Andrew Caines} \texttwemoji{orange}\texttwemoji{lemon} ~~~~
    {\bf Paula Buttery} \texttwemoji{orange}\texttwemoji{lemon} ~~~~
    {\bf Lisa Beinborn} \texttwemoji{green_apple} \\
    \texttwemoji{orange} Department of Computer Science \& Technology, University of Cambridge, U.K. \\
    \texttwemoji{lemon} ALTA Institute, University of Cambridge, U.K. \\
    \texttwemoji{green_apple} University of Göttingen, Germany \\ 
    \texttwemoji{orange} \texttt{firstname.secondname@cl.cam.ac.uk} \hspace{2mm}
    \texttwemoji{green_apple} \texttt{lisa.beinborn@uni-goettingen.de}
}

\begin{document}
\maketitle
\begin{abstract}
Language models strongly rely on frequency information because they maximize the likelihood of tokens during pre-training. As a consequence, language models tend not to generalize well to tokens that are seldom seen during training. Moreover, maximum likelihood training has been discovered to give rise to anisotropy: representations of tokens in a model tend to cluster tightly in a high-dimensional cone, rather than spreading out over their representational capacity. 

Our work introduces a method for quantifying the \textbf{frequency bias} of a language model by assessing sentence-level perplexity with respect to token-level frequency. We then present a method for reducing the frequency bias of a language model by inducing a syntactic prior over token representations during pre-training. Our \texttt{\textbf{Syntactic Smoothing}} method adjusts the maximum likelihood objective function to distribute the learning signal to syntactically similar tokens. This approach results in better performance on infrequent English tokens and a decrease in anisotropy. We empirically show that the degree of anisotropy in a model correlates with its frequency bias. 

\end{abstract}

\begin{tblr}{colspec = {Q[c,m]|X[l,m]}, stretch = 0}
    \cincludegraphics[width=1em, keepaspectratio]{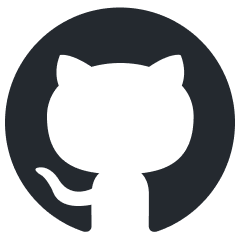} & {\footnotesize{ \href{https://github.com/rdiehlmartinez/syntactic-smoothing}{rdiehlmartinez/syntactic-smoothing}}}
\end{tblr}

\section{Introduction}

\begin{figure}[ht!]
    \centering
    \includegraphics[width=0.9\linewidth]{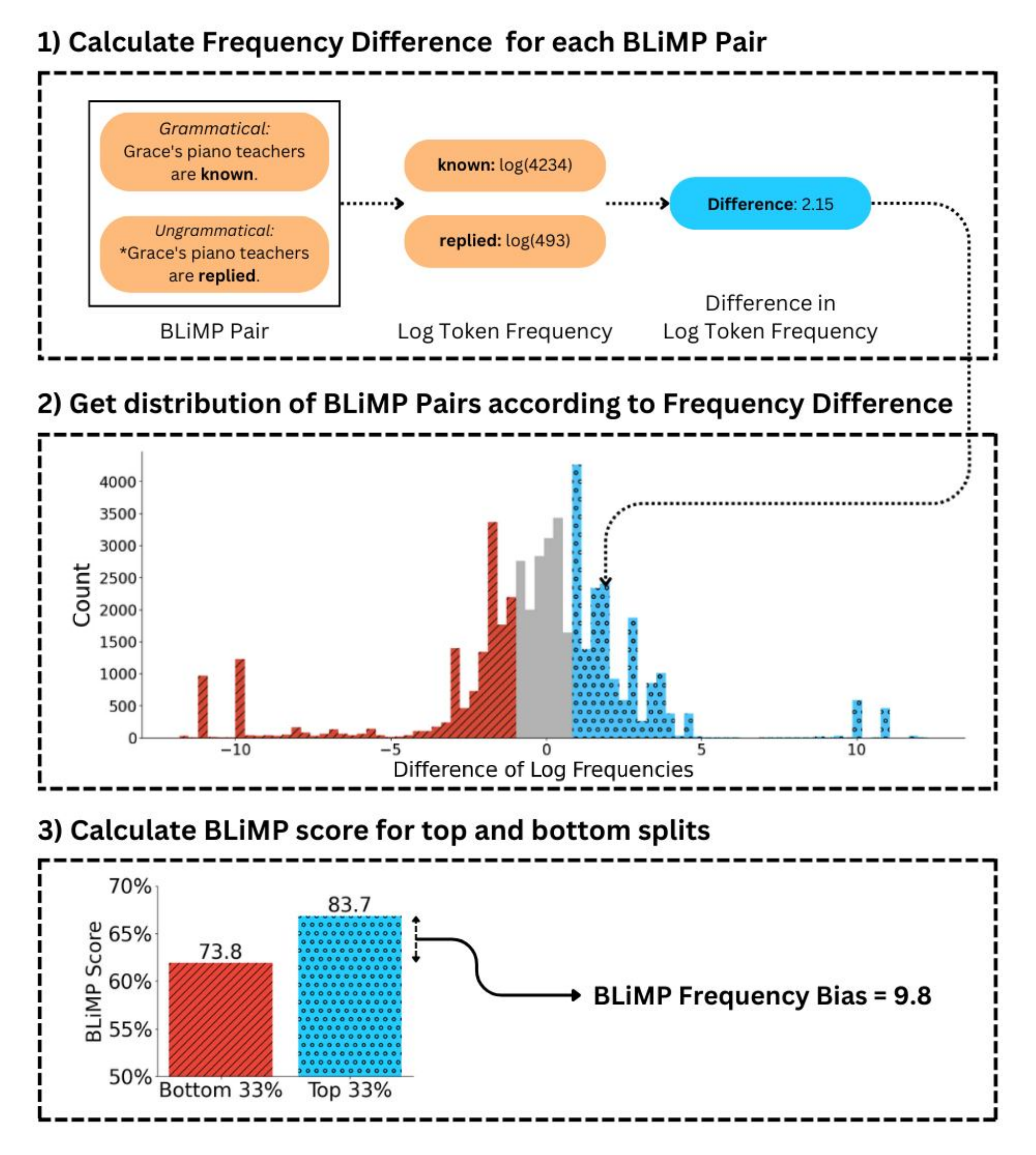}
    \caption{Illustration of the BLiMP \textbf{frequency bias} calculation used to evaluate a model's reliance on frequency statistics when making predictions. The example BLiMP values are from a baseline RoBERTa model.}
    \label{fig:blimp_bias}
    \vspace{-1em}
\end{figure}

Humans possess a remarkable ability to quickly understand the meaning of unknown words, given contextual cues. Consider the sentence, ``the Golden Gate Bridge has been \emph{obnebulated} every morning this week, limiting visibility of the Pacific Ocean.'' For many readers, `obnebulated' is probably not a familiar term, but we are likely to infer that 1) it is almost certainly a verb because of the -ed suffix and occurrence after perfective and passive auxiliaries, and 2) its meaning relates to visibility and climatic conditions.\footnote{ `Obnebulate' is an obsolete word meaning, ``To obscure with or as with a mist; to befog'' (\emph{Oxford English Dictionary}).} The ability to integrate unknown words based on syntactic and semantic context is essential for robust language understanding and still poses a significant challenge for language models. Nevertheless, Pre-trained Transformer Language Models (PLMs) have proven tremendously capable of solving a wide array of language processing tasks \citep{touvron2023llama, chowdhery2023palm}. 

Part of the success of PLMs can be attributed to the pre-training objective. Despite variations in architecture, the vast majority of language models are pre-trained to maximize the log-likelihood of a word, given the surrounding context \citep{devlin-etal-2019-bert, brown2020language, chowdhery2023palm, touvron2023llama}. As language use is characterized by a Zipfian distribution \citep{zipf}, language models are exposed to frequent tokens exponentially more often than infrequent ones during pre-training. Consequently, the representations of these frequent tokens are optimized based on exponentially more learning signals than those of low-frequency tokens. It has been shown that maximum likelihood objectives lead to representation degeneration in English language models because infrequent tokens are pushed into a narrow manifold of the representational space \citep{gao2018representation}. This representation degeneration problem is linked to the broader problem of \textbf{anisotropy}: the hidden states of a language model tend to cluster together into a small cone-shaped subspace, rather than over their full representational capacity \citep{arora2016latent, ethayarajh2019contextual, gao2018representation}. As language model evaluation is based on cumulative evaluation scores that conceal how well a model processes infrequent words, the disparities in the representational space are difficult to assess.  

Conventional language modeling approaches require large model sizes to effectively capture long-tail vocabulary distributions, limiting the scalability of these methods \citep{feldman2020does,haviv-etal-2023-understanding}. In this work, we propose \textbf{\texttt{Syntactic Smoothing}}: a syntactically-guided label smoothing approach to improve the representation of infrequent tokens in language models without resorting to perpetual increases of model and training data size. We smoothly distribute the backpropagation signal over syntactically similar tokens using a similarity metric based on part-of-speech (POS) tag distributions. Using this method, tokens that are seldom seen during training benefit from the more frequent updates of tokens that occur in similar syntactic functions. We evaluate our method using a new metric for quantifying the \textbf{frequency bias} of language models (illustrated in \cref{fig:blimp_bias}) and find that \texttt{Syntactic Smoothing} reduces both the frequency bias and the degree of anisotropy in a small English language model. We further explore the relationship between anisotropy and frequency bias and their effect on downstream performance. 

\vspace{-0.3em}
\section{Related Literature}

Through maximum likelihood training, language models implicitly learn to encode token frequency statistics. This training process gives rise to a frequency bias in models that constrains their ability to generalize to infrequent tokens. In this section, we begin by reviewing literature that discusses the challenges of generalizing linguistic knowledge to infrequent tokens. We then examine recent work that links the impact of token frequency to anisotropy in the models' representational space.

\subsection{Generalization to Infrequent Tokens}

Current approaches to language modeling rely heavily on the memorization of infrequent tokens to perform well on downstream tasks \citep{feldman2020does}. Recent analytical work has shown that certain layers of transformer models implicitly store memorized long-tail data \citep{haviv-etal-2023-understanding, kobayashi2023transformer}. \citet{feldman2020neural} demonstrate that models memorize atypical examples to achieve the highest accuracy on long-tailed data samples. This memorization hack, however, has only been shown to work well with over-parameterized models \citep{belkin2019reconciling}. While these studies present various metrics to evaluate memorization, these metrics do not capture how memorization impacts generalized linguistic understanding within the models. In our work, we address this gap by developing a metric that quantifies the extent of this frequency bias in relation to models' linguistic abilities.

Language use follows a Zipfian distribution, meaning that many tokens appear infrequently. Standard training objectives often require large models and noisy datasets with sufficient long-tail samples for effective generalization \citep{Zheng2022AnES}. However, improving generalization without excessive scaling can be achieved by training models with inductive priors that leverage linguistic information. On the lexical level, the integration of morphological and orthographic information during representation learning has been explored to obtain more fine-grained word embeddings \citep{salle-villavicencio-2018-incorporating, vulic-etal-2017-morph, cotterell-schutze-2015-morphological, bhatia-etal-2016-morphological, both-blunsom-compositional-morphology}. To improve syntactic generalization, the objective function has been enriched with auxiliary tasks, such as predicting constituency labels \citep{wang2023language}, hypernyms \citep{bai-etal-2022-better}, dependency tags \citep{cui2022lert}, and POS tags \citep{martinez-etal-2023-climb}. Some approaches have also shown promising results on rare word performance by constructing token embeddings that consider a word's surface form and surrounding context \citep{schick2019attentive, schick2020rare}.

\subsection{Anisotropy in Representational Space}
While frequency bias and generalization capabilities can be observed by analyzing model behavior on input--output patterns, representational analyses indicate that these phenomena are linked to the distribution of token representations. Language models trained as likelihood maximizers have been shown to yield degenerate representations for rare tokens \citep{gao2018representation}. Throughout training, infrequent tokens are disproportionately pushed in the negative direction of most hidden states, resulting in their clustering together irrespective of their semantic or syntactic properties. This clustering behavior leads to anisotropy: rather than occupying a large region of the representational space, token representations lie along a narrow manifold \citep{gao2018representation, ethayarajh2019contextual}. 

\subsubsection{Defining Anisotropy}

Anisotropy is defined as the inverse of isotropy: $1-I(v(\cdot))$. A representational space is isotropic if all the vector directions are distributed uniformly, meaning no particular direction is favored over another.

\citeauthor{arora2016latent}\ and \citeauthor{mu2018all} define isotropy as:

\begin{equation}
    I(v(\cdot)) \coloneq \frac{\min_{\norm{c}=1} Z(c)}{\max_{\norm{c}=1} Z(c)}
\end{equation}
where $c$ is a unit vector and $Z(c)$ is defined as the partition function over all tokens $w$ in the vocabulary $V$ , with representations $v(w)$:
$$
    Z(c) = \sum_{w \in V} \exp(c^Tv(w))
$$
In practice, this definition of isotropy is analytically infeasible to solve. In this paper, we follow an empirical approximation proposed by \citeauthor{ethayarajh2019contextual}: 
\begin{equation}
\label{eq:empirical-isotropy}
    I(v(\cdot)) \coloneq \mathbb{E}_{i\ne j}\big(1-\cos(v(w_i), v(w_j))\big)
\end{equation}
Here, $w_i$ and $w_j$ are two tokens sampled from the vocabulary, and $\cos$ is defined as taking the cosine similarity of the two word representations for $w_i$ and $w_j$.  

Despite its prevalence, the impact of anisotropy on a model's language understanding abilities remains unclear. Some studies suggest that reducing anisotropy improves performance on non-contextual benchmarks, sentence comparison tasks, and multilingual benchmarks \citep{bis-etal-2021-much, su2021whitening, rajaee2022isotropy}. Conversely, other research indicates that higher anisotropy might enhance semantic clustering tasks and that reducing anisotropy does not uniformly improve performance on common NLU tasks \citep{ait2023anisotropy, ding2022isotropy}. Furthermore, the relationship between anisotropy and maximum likelihood training has been questioned. Some researchers argue that isotropy exists in local manifolds of contextual word representations \citep{cai2020isotropy}, while others contend that anisotropy arises from the learning dynamics of the query and key attention matrices in transformer models \citep{godey2024anisotropy}.

\subsubsection{Reducing Anisotropy}

Existing methods to reduce anisotropy broadly fall into three categories. The first group of approaches transforms the hidden states of language models to remove semantically uninformative directions and to preserve the dimensions of maximal isotropy \citep{arora2016simple, mu2018all, raunak2019effective, su2021whitening, bis-etal-2021-much}. This intervention style is based on the assumption that the top singular dimensions of pre-trained word representations encode frequency statistics rather than semantic or lexical information \citep{mu2018all}. The second category of methods introduces novel training objectives and regularization terms that reduce the effects of anisotropy \citep{gong2018frage, gao2018representation, wang2019improving}. This type of approach places an inductive bias on representations that push the embeddings of frequent and infrequent words to occupy a similar semantic space. The third set of approaches explores different training paradigms to directly minimize anisotropy, such as using normalizing flow models \citep{li2020sentence} or manipulating the gradients used in maximum likelihood models \citep{yu2022rare}

\vspace{1em}

While frequency bias and anisotropy are prevalent in language modeling, quantifying their effects and understanding their impact on generalization, particularly for infrequent words, remains an open area of research. Our paper introduces a novel method for improving the representation of infrequent tokens by integrating linguistic information. Moreover, we hypothesize that adjusting the learning process to better represent infrequent tokens will also reduce anisotropy, as these two phenomena are interconnected.

\section{Frequency Bias}
\label{section:freq-bias}

We investigate frequency effects using a zero-shot test of grammatical capability known as BLiMP: The Benchmark of Linguistic Minimal Pairs \cite{warstadt2020blimp}. BLiMP comprises 67 datasets (or ``subtasks''), each consisting of 1,000 pairs of grammatical and ungrammatical sentences that differ only with respect to a specific linguistic characteristic (covering syntax, morphology, and semantics). Language models are tasked with assigning a higher likelihood to the grammatical sentence. The grammatical generalization capabilities of a language model are often summarized by averaging the accuracies achieved across the 67 BLiMP tasks. While random guessing scores 0.5, state-of-the-art models have achieved scores of 0.87 when trained on large datasets, and models trained on the 10M-word BabyLM dataset have achieved scores up to 0.80 \citep{warstadt2023findings}. 

BLiMP is carefully balanced to ensure individual tokens occur equally in both sentence types. However, within a single pair, there may be an imbalance in average token frequency: For instance, the sentence
\textit{Grace's piano teachers are \textbf{known}} has a log frequency of 8.35 while its associated minimal pair \textit{Grace's piano teachers are \textbf{replied}} has a log frequency of 6.20.  We hypothesize that despite the minimal difference in BLiMP pairs, models trained in a typical manner will be biased by token frequency when determining grammatical acceptability.

Our goal is to quantify how language model performance differs between BLIMP pairs with large positive frequency differences (where the correct sentence has more frequently occurring tokens) and with large negative frequency differences (where the correct sentence has much less frequently occurring tokens). We do so in two steps.

First, for each BLIMP sentence pair, we calculate the average (natural log) frequency of the differing tokens. Frequencies of individual tokens are computed with respect to a model's training data; for instance, in the example above the token \textit{\textbf{known}} has a log frequency of 8.35 in the training data. Sentence pairs are then ranked by the relative difference in these average frequencies, where positive values indicate a higher average frequency for the acceptable sentence. These relative differences form a distribution, as shown in the middle plot of \cref{fig:blimp_bias}. 

Then, we compute the BLiMP score using pseudo log-likelihood \citep{salazar2020masked} for BLIMP pairs in the upper and lower thirds of the relative frequency difference distribution. We exclude the middle third, as these represent pairs with minimal frequency differences (see the frequency plot for details). We define a model's \textbf{frequency bias} as the difference between the two BLiMP scores. The entire process is illustrated in \cref{fig:blimp_bias}. 

In practice, we find that standard transformer language models, such as OPT-125M \citep{zhang2022opt}, RoBERTa-base \citep{liu2019roberta}, and T5-base \citep{raffel2020exploring}, exhibit a frequency bias as high as 13.7\%. Our goal is to develop a model that can attain a frequency bias close to zero while attaining a high BLiMP score: that is, a model that makes determinations on the grammatical acceptability of sentences based solely on relevant linguistic aspects, rather than relying on possibly misleading statistical artifacts of the training data. 

\section{Syntactic Smoothing}

We hypothesize that transformer language models exhibit a strong frequency bias due to their maximum likelihood training objective, which limits infrequent tokens from receiving useful learning signals and thus hinders their ability to effectively encode linguistic information. To address this, we propose at each learning step to backpropagate the learning signal of a target token to all other tokens serving similar syntactic roles; this benefits infrequent tokens that appear less often in the training data.

\textbf{\texttt{Syntactic Smoothing}} implements this strategy by distributing a portion of every update signal to all syntactically similar tokens using a syntactic similarity metric (operationalized below). This results in the representation of infrequent tokens approaching the average representation of all tokens that serve a similar syntactic function; e.g., the representation of a niche word like `obnebulated' would encode its syntactic role as a verb.

Our method consists of two components; (1) a similarity metric that uses part-of-speech distributions as a coarse proxy for syntactic similarity, and (2) an adjustment to the loss function to smooth the backpropagation signal over syntactically similar tokens during pre-training. 

\subsection{Syntactic Similarity Score}\label{sec:sim}

The syntactic similarity between two tokens can be measured in multiple ways, e.g., by using surface features, dependency labels, or even the predictions of a teacher language model \citep{hinton2015distilling}. Here, we present a simple measure that acts as a coarse approximation for syntactic similarity: we consider two tokens to be similar if they have a similar distribution of part-of-speech tags in the training set.

We evaluate the syntactic similarity between tokens prior to training, as a one-off preprocessing step over the entire training set. First, we use the part-of-speech (POS) tagger from the NLTK package \citep{bird2009natural} to assign each word in the training set to one of 12 universal POS tags, based on its given context \citep{petrov-etal-2012-universal}.\footnote{The 12 tags in the NLTK tagger are given here: \url{https://www.nltk.org/book/ch05.html\#tab-universal-tagset}. They are derived from the 17 tags in the Universal Dependencies tagset.} We then tokenize the training data into sub-word tokens and assign each token the POS tag corresponding to the word it belongs to in each instance. As words can take on a different part of speech depending on the context, we count the number of times each token in our vocabulary $V$ appears as each POS tag in the training data, producing a 12-valued vector. This results in a matrix $M \in \mathbb{R}^{|V|\times 12}$ containing the distribution over POS tags for each token. Finally, we can compute the similarity of two tokens $V_i$ and $V_j$ using the cosine similarity of their POS distributions: $$ \text{Syntactic Similarity(i, j)} = \frac{M_i^TM_j}{||M_i|| \cdot ||M_j||}$$

Note that while in this paper we define syntactic similarity via cosine similarity, any real-valued distance metric or divergence can be used. The similarity function does not need to be symmetric, although we note that symmetric functions provide computational advantages as only half the values need to be computed and stored. Also, note that our methodology does not depend on a specific choice of POS tagger.

\begin{figure}[t]
    \centering
    \includegraphics[width=\linewidth]{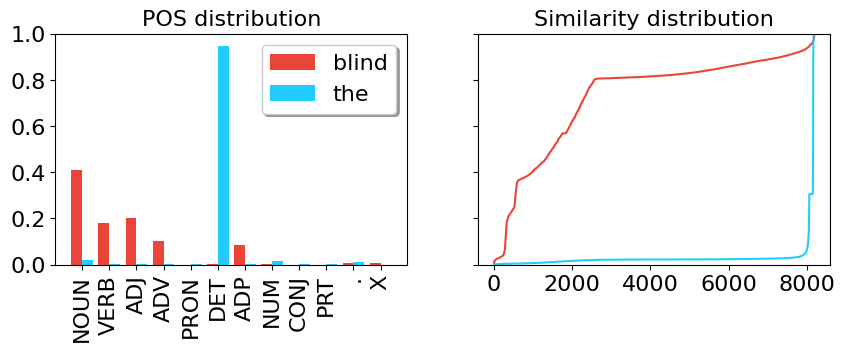}
    \caption{Part-of-speech distributions and similarity distributions for the subword tokens ``blind'' and ``the''. Similarities are computed as cosine-similarities against every other token in the vocabulary and sorted.}
    \label{fig:distributions}
    \vspace{-1em}
\end{figure}

We provide the POS distributions and similarity distributions for the example tokens ``blind'' and ``the'' in \cref{fig:distributions}. Notice that ``the'' occurs almost exclusively as a determiner and is not similar to many other tokens, whereas ``blind'' occurs as a noun, verb, adjective, and adverb and has a high similarity to more than half the other tokens in the vocabulary.

\subsection{Smoothing the Backpropagation Signal}\label{section:smoothing}
Modern pre-training objectives implement likelihood maximization using a cross-entropy loss between the label of the correct word and predicted probabilities from a forward pass of the model. \texttt{Syntactic Smoothing} makes a small adjustment. Instead of a one-hot encoding, the target vector $t$ becomes a distribution across the entire vocabulary with some of the signal on the correct label $j$ and the rest of the signal distributed across all other tokens $i$ according to the syntactic similarity metric used:
\begin{equation}
\label{eq:signal-distribution}
    t_i=\left\{
  \begin{array}{@{}ll@{}}
    (1-\alpha), & \text{if}\ i=j \\
    \frac{s(i,j)}{\sum_{k=0}^{|V|}{s(i,k)}} \times \alpha & \text{otherwise}
  \end{array}\right.
\end{equation}

\noindent
where $\alpha$, the smoothing parameter, determines the proportion of the error signal reserved for the correct word and $s$ is our part-of-speech similarity metric. We experiment with different values for $\alpha$, noting that $\alpha=0$ is the standard likelihood maximization task. We also investigate the use of a pacing function that linearly decreases $\alpha$ so that at the start of training the majority of the signal is propagated to other syntactically similar tokens and by the end of training nearly all of the error signal is sent to the correct token to ensure that the model still optimizes perplexity. 

In practice, we also find it beneficial to apply a temperature scaling function to the syntactic similarity distribution. Thus, rather than using the raw syntactic similarity scores, $s(i,j)$, in \cref{eq:signal-distribution}, we use the temperature-scaled similarity scores:

$$
s'(i,j) = \frac{\exp\left(\frac{s(i,j)}{\tau}\right)}{\sum_{k=1}^{|V|} \exp\left(\frac{s(i,k)}{\tau}\right)}
$$
where $\tau$ defines the temperature which we set to $\tau=0.025$. 


\subsection{Experimental Setup}
\label{subsection:experimental_setup}

Our experiments focus on smaller language models and datasets due to computational constraints and the particular challenges of generalizing to uncommon instances under resource-constrained training conditions \citep{warstadt2023findings, martinez-etal-2023-climb}. 

\paragraph{Data} \label{paragraph:data} We use the dataset published as training data for the BabyLM challenge at the 2023 CoNLL workshop \citep{warstadt2023findings}. It contains roughly 10 million tokens sampled from pre-existing datasets, covering a wide range of domains including transcribed speech (both adult-directed and child-directed), movie subtitles, Wikipedia articles, and books. The dataset was constructed to be similar to the input received by children --- 56\% comes from transcribed speech and 40\% comes from sources intended for children. 

\paragraph{Model} We use a small 8-layer encoder-style RoBERTa model with pre-layer normalization \cite{huebner-etal-2021-babyberta}. We report the hyper-parameter settings we use throughout all experiments in \cref{tbl:appendix_hyperparams} (\cref{section:appendix-hyperparameters}) and computational requirements in \cref{section:appendix-computational-requirements}. We use a BPE tokenizer \citep{sennrich-etal-2016-neural} with a vocabulary size of 8192 as recommended in previous work \cite{martinez-etal-2023-climb}. 

\paragraph{Evaluation} We evaluate the BLiMP frequency bias of our models, as defined in \cref{section:freq-bias}, on the evaluation set of BLiMP. To compute anisotropy we use the formulation defined in \cref{eq:empirical-isotropy}; We sample 1,000 pairs of random word tokens with their surrounding context from the training set, and compute the cosine similarity of their hidden representation at each of the 8 layers of the RoBERTa model. To obtain a model's final anisotropy value, we average the anisotropy scores across the 8 layers. Additionally, we finetune and evaluate each model on two downstream sentence-level tasks, COLA \citep{warstadt-etal-2019-neural} and SST-2 \citep{socher-etal-2013-recursive}, as well as two language inference tasks, MNLI \citep{williams-etal-2018-broad} and QNLI \citep{rajpurkar-etal-2016-squad, wang-etal-2018-glue}.

\paragraph{Baselines}

We introduce three types of baselines: 
\begin{enumerate}
    \item \textbf{Popular open-source transformer models}: OPT-125M \citep{zhang2022opt}, RoBERTa-base \citep{liu2019roberta}, and T5-base \citep{raffel2020exploring}, pre-trained from scratch on the same dataset we describe in \cref{subsection:experimental_setup}. We use the default configuration for each model resulting in a varied number of parameters.
    \item \textbf{Base Model}: The small RoBERTa model described above without \texttt{Syntactic Smoothing}.
    \item \textbf{Label Smoothing}: The base model trained with label smoothing \citep{szegedy2016rethinking}.  We train a baseline with a low-level of smoothing ($\alpha=0.2$) and a mid-level of smoothing ($\alpha=0.5$). Note that \texttt{Syntactic Smoothing} can be seen as a linguistically-guided version of the standard label smoothing approach, in which the learning signal is distributed to all tokens uniformly.
\end{enumerate}

\paragraph{Our Models} We train our models with \texttt{Syntactic Smoothing} using the same two $\alpha$ values as the label smoothing baselines to facilitate comparison. We also run variants using the linear pacing function presented in \cref{section:smoothing} which linearly decreases the smoothing from an initial value of $\alpha$ to zero across training. For these variants, we use the same two values of smoothing, as well as an additional high value of $\alpha=0.8$ giving a total of five \texttt{Syntactic Smoothing} variants.\footnote{We do not include unpaced \texttt{Syntactic Smoothing} with a high value of $\alpha$ as initial experiments found that distributing such a high proportion of the learning signal away from the correct token leads to high perplexity and poor downstream performance.}

\section{Results}
\label{sec:results}

\begin{table*}[ht!]
\centering
\small
\begin{tabular}{ll||cc|ccccc}
\toprule
\textbf{Model}  & $\alpha$ & \textbf{Bias}  & \textbf{Anisotropy} & \textbf{BLiMP} & \textbf{COLA} & \textbf{SST-2} & \textbf{MNLI} & \textbf{QNLI}  \\
\midrule
Base Model & -&9.8 & 51.3 & 71.4 & 71.4 & 82.9 & 69.6 & 79.7 \\
\midrule
\multirow{2}{*}{Label Smoothing} &Low & 5.5 & 40.2 & 73.2 & 70.7 & 84.0 & \textbf{70.1} & 80.0 \\
&Mid & 2.7  & 40.3 & 73.0 & 71.5 & 82.2 & 69.0 & 79.4 \\
\midrule
\multirow{5}{*}{\texttt{Syntactic Smoothing}}&Low  & 2.9 & 39.7 & \textbf{73.2} & 70.7 & 84.9 & 69.7 & 79.2 \\
&Mid  & \textbf{-0.2} & 33.8 & 72.1 & \textbf{71.9} & 83.5 & 67.2 & 79.4 \\
&Paced Low & 7.4 & 39.9 & 71.9 & 70.5 & \textbf{85.2} & 70.0 & \textbf{80.4}\\ 
&Paced Mid& 5.7 & 34.5 & 72.3 & 71.8 & 84.0 & 68.2 & 78.9\\ 
&Paced High & 5.2 & \textbf{31.0} & 72.2 & 70.5 & 83.7 & 67.7 & 79.1 \\ 
\bottomrule
\end{tabular}
\caption{\label{tbl:full-results} We report bias~($\downarrow$), anisotropy~($\downarrow$), BLiMP~($\uparrow$) score, and accuracy or correlation scores ($\uparrow$) on two downstream sentence-level tasks -- COLA and SST-2 -- and two downstream language inference tasks -- MNLI and QNLI -- for our MLM baseline, two label smoothing (LS) baselines, and five \texttt{Syntactic Smoothing} (\texttt{SyS}) variants. \texttt{SyS}-P variants use linear pacing to reduce the smoothing factor to zero over training.}
\end{table*}

Our results are summarized in \cref{tbl:full-results}. We find that our method reduces frequency bias while retaining strong language modeling capabilities. At the same time, we observe that the models with the lowest frequency bias also demonstrate the lowest anisotropy. We then extend our analysis beyond the specific phenomenon of frequency bias and anisotropy by examining the impact of \texttt{Syntactic Smoothing} on the linguistic generalization capabilities of the model and its downstream performance after finetuning. Finally, we find that an alternative syntactic scoring metric leads to similar results as the cosine-based definition.

\subsection{Anisotropy and Frequency Bias}
We conduct analyses to inspect the learning dynamics of our method and its effect on frequency bias and anisotropy in more detail. 
\begin{figure}[h]
    \centering
    \includegraphics[width=0.90\linewidth]{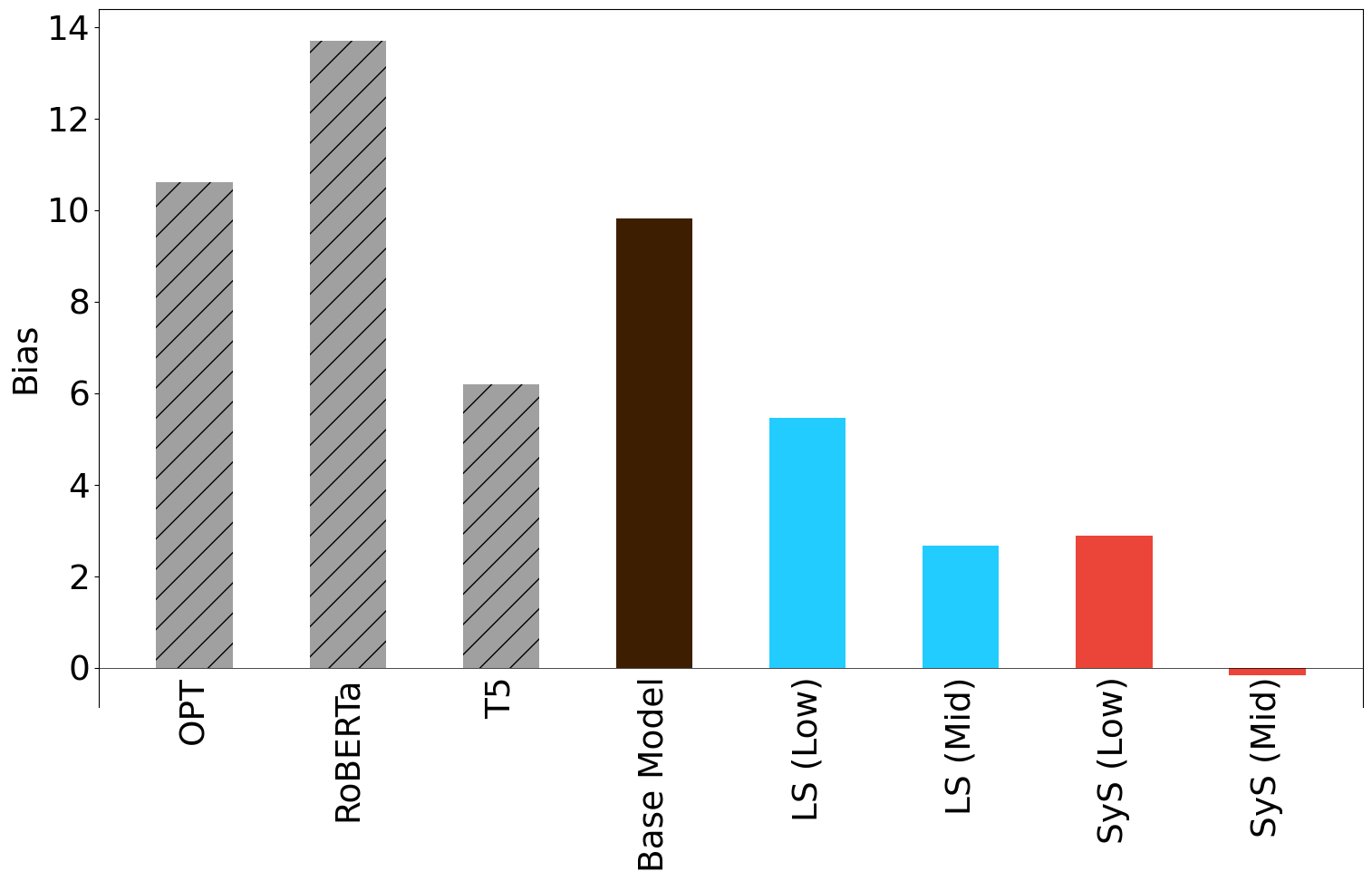}
    \caption{Frequency bias plotted for the three open source pre-trained models, our base model, the two label smoothing (LS) baselines and our two \texttt{Syntactic Smoothing} (\texttt{SyS}) models.}
    \label{fig:biases}
    \vspace{-1em}
\end{figure}

\paragraph{\texttt{Syntactic Smoothing} reduces frequency bias.}
We find that all four pre-trained models exhibit strong frequency bias (see \cref{fig:biases}); they are more likely to incorrectly prefer ungrammatical sentences if they contain tokens that occur more frequently during training. This confirms our hypothesis that the evaluation of generalization capabilities is obfuscated by frequency effects. 

By contrast, the two \texttt{Syntactic Smoothing} variants successfully reduce the frequency bias. The frequency bias is almost completely removed in the case of the \texttt{Mid} variant, which distributes exactly half of the training signal to syntactically similar tokens. We further observe that the Label Smoothing baselines also reduce bias but to a lesser extent than the corresponding \texttt{Syntactic Smoothing} models with the same degree of smoothing. 

\begin{figure}[h!]
    \centering
    \includegraphics[width=\linewidth]{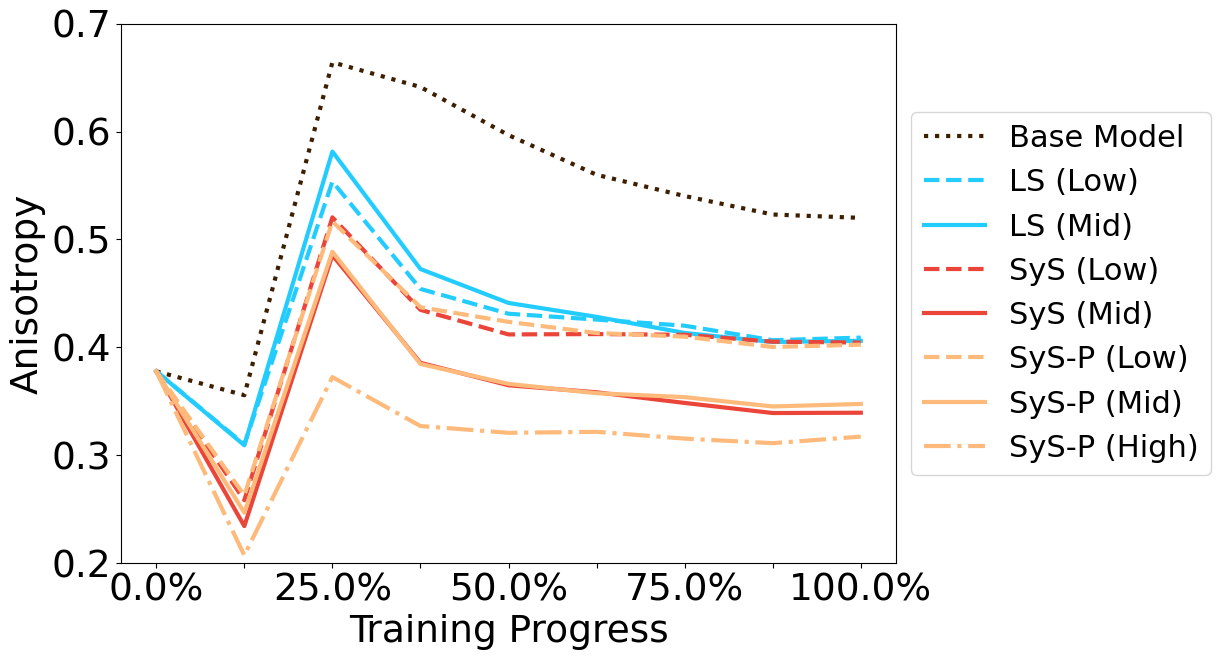}
    \caption{Anisotropy learning dynamics plotted for the baseline RoBERTa model, the two label smoothing (LS) baselines and our \texttt{Syntactic Smoothing} (\texttt{SyS}) models. Values in parentheses indicate the degree of smoothing.}
    \label{fig:anisotropy-learing-dynamics}
    \vspace{-1em}
\end{figure}

\paragraph{\texttt{Syntactic Smoothing} reduces anisotropy.}

As shown in \cref{tbl:full-results}, \texttt{Syntactic Smoothing} reduces anisotropy over both the base model and label smoothing baselines.\footnote{Note that we do not compute the anisotropy for the three open-source pre-trained models (OPT, RoBERTa, T5) because these models use different architectural configurations than the models we train (e.g., larger hidden dimensions).} Label smoothing reduces anisotropy, but not to the same extent as our \texttt{Syntactic Smoothing} models. To better understand how anisotropy develops in a model, we compute the model's anisotropy scores at eight checkpoints during training, as shown in \cref{fig:anisotropy-learing-dynamics}. We find that a greater degree of smoothing leads to a greater reduction in anisotropy for our \texttt{Syntactic Smoothing} variants (it is less clear if this is the case for label smoothing), supporting our hypothesis that syntactic initialization helps promote better representation learning across the model's vocabulary. We also find that the pacing method leads to lower anisotropy than the flat method, with \texttt{SyS}-P (High) achieving the lowest anisotropy throughout.

\begin{figure}[h]
    \centering
    \includegraphics[width=\linewidth]{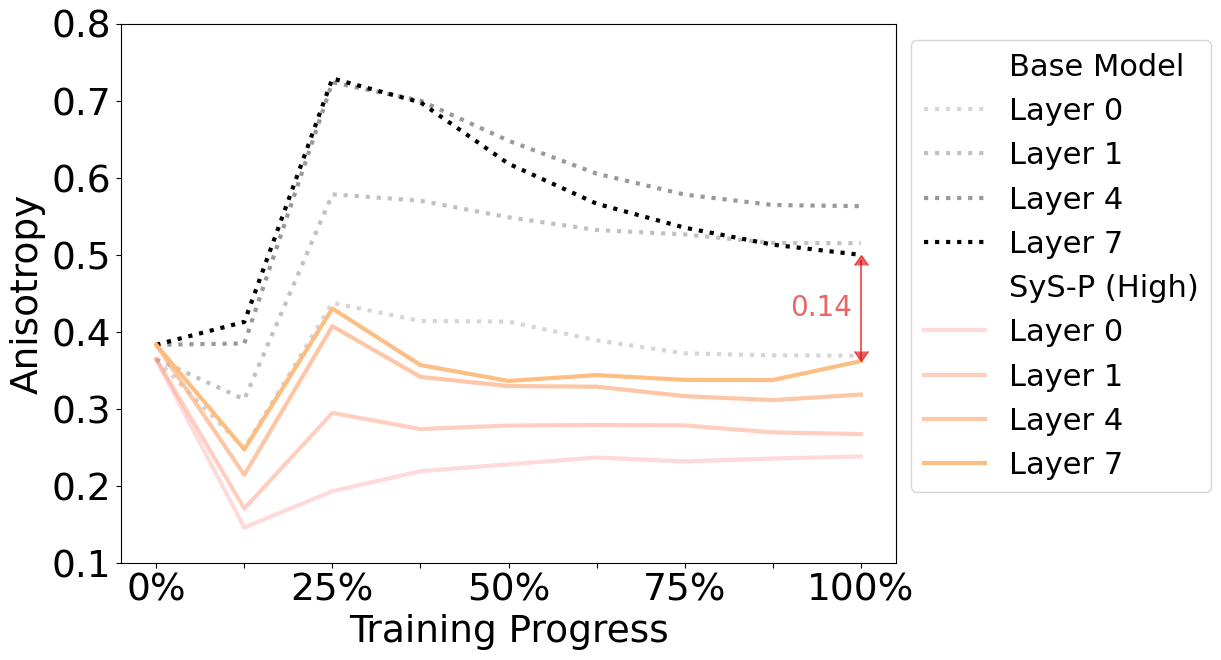}
    \caption{Anisotropy learning dynamics plotted for the baseline model and the paced \texttt{Syntactic Smoothing} model with high smoothing, across some of the models' layers. We highlight the difference in anisotropy of the final layer across the two models at the end of training.}
    \label{fig:anisotropy-layers}
    \vspace{-1em}
\end{figure}

Over the course of training, we observe a consistent double-dip trend: an initial dip followed by a sudden rise, followed by a second slow decrease in anisotropy. The \texttt{Syntactic Smoothing} models do not see as large a sudden rise, maintaining a lower anisotropy throughout. To examine the learning dynamics in more detail, we also plot the evolution of the anisotropy across several layers of our baseline model and the \texttt{SyS}-P (High) variant, given in \cref{fig:anisotropy-layers}. Two observations stand out. The anisotropy of all layers in the \texttt{Syntactic Smoothing} model is lower than in the corresponding layers in the baseline model across the entire learning process. In both the baseline model and the \texttt{Syntactic Smoothing} model, earlier layers have lower anisotropy; this finding agrees with the same observation made by \citeauthor{ethayarajh2019contextual}. Notably, in the final layer—commonly used for sentence representations in downstream tasks—the anisotropy of the \texttt{Syntactic Smoothing} model remains consistently low and does not increase significantly during training, in contrast to the drastic fluctuation observed in the baseline model. 

\paragraph{Frequency bias and anisotropy are correlated.}

\begin{figure}[h]
    \centering
    \includegraphics[width=\linewidth]{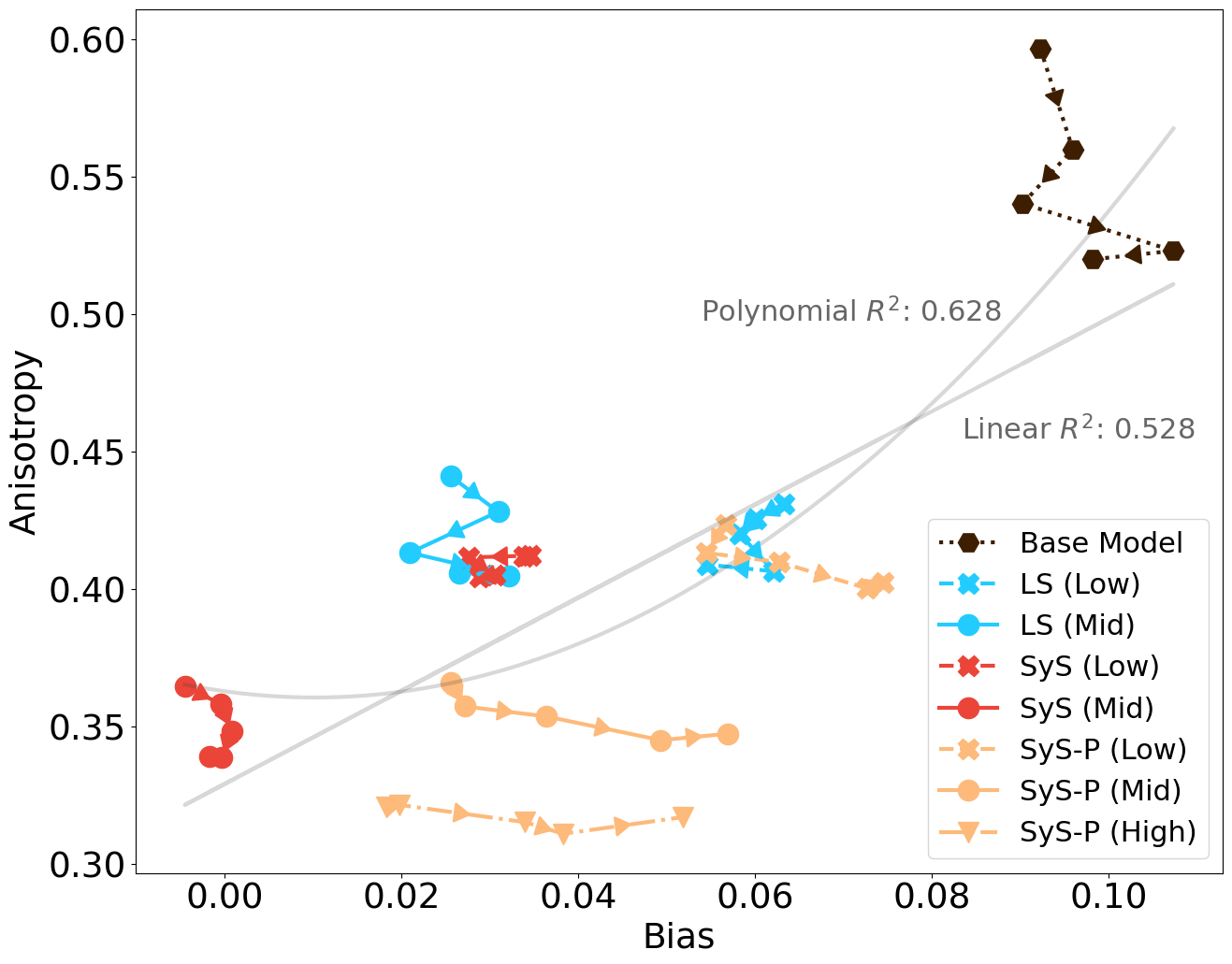}
    \caption{Pairs of anisotropy, and frequency bias for the baseline RoBERTa model, the two label smoothing baselines and our \texttt{Syntactic Smoothing} models. The arrows indicate increasing training progress (starting after 50\% of training has completed).}
    \label{fig:bias-anisotropy-correlation}

\end{figure}

For each model, we compute the model's frequency bias and anisotropy at multiple training stages. We plot the learning dynamics of anisotropy and frequency bias in \cref{fig:bias-anisotropy-correlation}, only including the points after 50\% of training has been completed to avoid the noisy first dip observed in the anisotropy dynamics above. We find a positive Pearson correlation of 0.73 and a polynomial goodness-of-fit $R^2$ score of 0.63 between these two metrics.

It is also evident that the pacing approach re-introduces frequency bias towards the end of training, as the degree of smoothing is linearly reduced to zero. It is noteworthy that the final anisotropy and bias are lower than the baseline model, and completing training without any smoothing may be beneficial for downstream tasks, as explored in the next section.

\subsection{Effects of Smoothing on Downstream Tasks}
While our method primarily aims to enhance the representation of infrequent tokens, we sought to investigate the potential for improvement in standard evaluation measures, given the limited number of affected test instances. Nonetheless, we observe that all the \texttt{Syntactic Smoothing} models, as well as the label smoothing models, achieve better BLIMP scores than our baseline model (see \cref{tbl:full-results}). These results suggest that methods that smooth label distributions, whether through a syntactic prior or a simpler uniform smoothing approach, enhance the representation of all tokens, including the more frequent ones.

We had concerns that softening the frequency bias with our method might lead to degraded performance in downstream tasks for which frequency can be a strong proxy. As a control condition, we finetune our model on two sentence-level tasks (COLA, SST-2) and two language inference tasks (MNLI and QNLI), both of which are part of the GLUE \citep{wang-etal-2018-glue} benchmark. We find that none of the \texttt{Syntactic Smoothing} objectives result in substantial performance degradation on these NLU tasks (see the last four columns of Table~\cref{tbl:full-results}), and in fact note that for some tasks, such as SST-2, the \texttt{Syntactic Smoothing} models yield uniform increases in performance.\footnote{While not comparable apples-to-apples, we report NLU performance for the open-source baselines in \cref{section:opensource-baselines-nlu-results}.}   

\subsection{Alternative Measures of Syntactic Similarity}

In \cref{sec:sim} we define the syntactic similarity score that is used by the \texttt{Syntactic Smoothing} approach as the cosine similarity between POS distributions. To examine how this specific choice of similarity metric impacts our approach, we replace the cosine-based definition with a Jensen Shannon-based definition:
$$ \frac{1}{2}\big[ \text{KL}(M_i, M_j ) + \text{KL}(M_j, M_i)\big],$$
where KL$(M_i, M_j)$ is the Kullback-Leibler divergence between the POS distributions, $M_i$ and $M_j$, for the vocabulary items $V_i$ and $V_j$. 

\begin{table}[t]
\centering
\small
\begin{tabular}{l||cc|ccccc}
\toprule
\textbf{Model}  &  \textbf{Bias}  & \textbf{Anisotropy} & \textbf{BLiMP} \\
\midrule
Base Model & 9.8 & 51.3 & 71.4  \\
\midrule
\texttt{SyS} (Mid) \hspace{0.42cm} [JS]  & 3.6 & 34.7 & 71.3 \\
\texttt{SyS} (Low) \hspace{0.38cm} [JS]  & 4.1 & 34.6 & 73.3  \\
\texttt{SyS}-P (High) \hspace{0.05cm} [JS] & 6.6 & 36.7  & 72.5  \\ 
\texttt{SyS}-P (Mid) \hspace{0.15cm} [JS] & 8.4 & 39.1 &  73.0 \\ 
\texttt{SyS}-P (Low) \hspace{0.12cm} [JS] & 5.0 &  34.5 & 72.9 \\ 
\bottomrule
\end{tabular}
\caption{\label{tbl:jsd-similarity-metric-results}
Results for bias~($\downarrow$), anisotropy~($\downarrow$), and BLiMP~($\uparrow$) score for \texttt{Syntactic Smoothing} (\texttt{SyS}) models that use a Jensen Shannon-based [JS] definition of the similarity metric.}
\end{table}

Summarized in \cref{tbl:jsd-similarity-metric-results}, we note that the effect of using a Jensen Shannon-based definition of the similarity metric yields a similar (albeit slightly smaller) decrease in frequency bias and anisotropy, as compared to the standard cosine-based definition of the similarity metric.   

\section{Conclusion}

Our work studies the phenomenon of \textbf{frequency bias} in language models that degrades the performance of these models on tokens infrequently observed during training. We develop a novel method for quantifying the degree to which a language model prefers grammatically incorrect sentences that contain frequent tokens over grammatically correct sentences containing infrequent tokens.  We introduce a new training approach, \texttt{Syntactic Smoothing}, that distributes the backpropagation signal to syntactically similar tokens. Using a coarse approximation of syntactic similarity based on part-of-speech tags, we show that this approach can remove the frequency bias without degrading broader language understanding. We also find that reductions in frequency bias are strongly correlated with reductions in a model's anisotropy. Our findings provide a novel angle through which to observe the role of anisotropy in language modeling.

\section*{Ethical Impact}

Studying long-tail data comes with some known ethical concerns. Previous research has found that names of female and non-white persons tend to fall in the long-tail of many datasets which can result in less efficient neural representations of these names compared to names of male and white persons\citep{wolfe-caliskan-2021-low}. Our paper does not directly study whether the methods we develop affect these implicit biases, although we would suspect that our approach might help remove some of these biases (without further experimentation this, however, remains a risk of our work). 

Along similar lines, we also do not conduct a thorough analysis to determine whether the curated BabyLM training set we use contains offensive data or uniquely identifies individuals. For an overview of the pre-processing steps that were done to remove harmful data from the BabyLM corpora, we refer the reader to the BabyLM proceedings \citep{warstadt2023findings}.

We also note that the use of large-scale black-box LLMs makes studying infrequent token representations and their downstream effects more difficult. Our use of smaller LMs helps increase transparency and facilitates the reproducibility of our method by research groups with small computational budgets.

\section*{Limitations}
Our methods use English-only data, and thus assume an English-centric notion of word functions. For the syntactic information, we use the POS tags provided by the NLTK tagger. As this tagger was trained on a separate dataset, this may suggest our method relies on additional data in order to best represent infrequent words. However, in initial experiments with an unsupervised tagger trained only on the 10M-word dataset, we achieved similar results. Additionally, the models we experiment with are all relatively small and, while we assume that our results can be scaled up to larger architectures, our limited computational resources do not allow us to collect empirical evidence. In future work, we plan to further explore the impact of \texttt{Syntactic Smoothing} on models with autoregressive architectures and larger training datasets. We also hope future work will apply our method to more languages, possibly leveraging unsupervised POS taggers for these languages, and evaluate the effect of \texttt{Syntactic Smoothing} on different downstream tasks (particularly tasks with irregular vocabulary frequency distributions). 

\section*{Acknowledgements}
The experiments reported in this paper were performed using resources provided by the Cambridge Service for Data Driven Discovery (CSD3) operated by the University of Cambridge Research Computing Service, provided by Dell EMC and Intel using Tier-2 funding from the Engineering and Physical Sciences Research Council (capital grant EP/T022159/1), and DiRAC funding from the Science and Technology Facilities Council. Richard Diehl Martinez is supported by the Gates Cambridge Trust (grant OPP1144 from the Bill \& Melinda Gates Foundation). Z\'ebulon Goriely's work is supported by The Cambridge Trust. Lisa Beinborn's work is partially supported by the Dutch National Science Organisation (NWO) through the VENI program (Vl.Veni.211C.039).
Andrew Caines and Paula Buttery are supported by Cambridge University Press \& Assessment.

\bibliography{anthology,custom}
\bibliographystyle{acl_natbib}

\appendix

\section{Experimental Hyperparameters}
\label{section:appendix-hyperparameters}
\begin{table}[h!]
    \centering
    \small
    \begin{tabular}{l|r}
        \toprule
             Parameter & Value \\
        \midrule
             Layer Norm EPS& 1e-5 \\
             Learning Rate & 0.001 \\
             Optimizer & AdamW \\
             Scheduler Type & Linear\\
             Max Steps & 200,000 \\
             Warm-up Steps & 50,000 \\
             Total Batch Size & 512 \\
             Vocab Size & 8192 \\
             Hidden Dimension Size & 256 \\
             Max. Sequence Length & 128 \\
             Num. Attention Layers & 8 \\
             Num. Attention Heads & 8 \\
             Model Architecture & RoBERTa (Pre-LN) \\
        \bottomrule
    \end{tabular}
    \caption{Hyperparameter settings which are constant across all experiments}
    \label{tbl:appendix_hyperparams}
\end{table}

These hyperparameters are taken from \citet{martinez-etal-2023-climb} who tuned the RoBERTa model for the 10M-word BabyLM dataset. 

\section{Computational Requirements}
\label{section:appendix-computational-requirements}

We purposefully train a small-scale LM for our experiments. The total amount of the trainable parameters in our model is \textbf{12,750,336}. Each of our experiments trains for approximately 14-20 GPU hours, using a server with one NVIDIA A100 80GB PCIe GPU, 32 CPUs, and 32 GB of RAM for all experiments. Below, we report a subset of the output of the \emph{lscpu} command:

\begin{tcolorbox}[left=5pt,right=5pt,top=5pt,bottom=5pt]
\small
\begin{verbatim}
Architecture:        x86_64
CPU op-mode(s):      32-bit, 64-bit
Address sizes:       46 bits physical, 
                     48 bits virtual
Byte Order:          Little Endian
CPU(s):              32
On-line CPU(s) list: 0-31
Vendor ID:           GenuineIntel
Model name:          Intel(R) Xeon(R)
                     Silver 4210R CPU
                     @ 2.40GHz
CPU family:          6
Model:               85
Thread(s) per core:  1
Core(s) per socket:  1
Socket(s):           8
Stepping:            7
BogoMIPS:            4800.11
\end{verbatim}
\end{tcolorbox}
\vfill
\section{Word Class Versus Word Frequency Analysis}
\label{section:word-class-versus-word-frequency}

Broadly, we find that content words, primarily nouns, are over-represented in low-frequency tokens. We moreover, find that the syntactic distribution across POS tags changes considerably when comparing the top 100 and bottom 100 most and least frequently occurring tokens. This analysis suggests that poor performance on infrequent tokens has a particularly strong effect on a model's inability to correctly model specialized noun vocabulary items. 

\begin{figure}[h!]
    \centering
    \includegraphics[height=4cm]{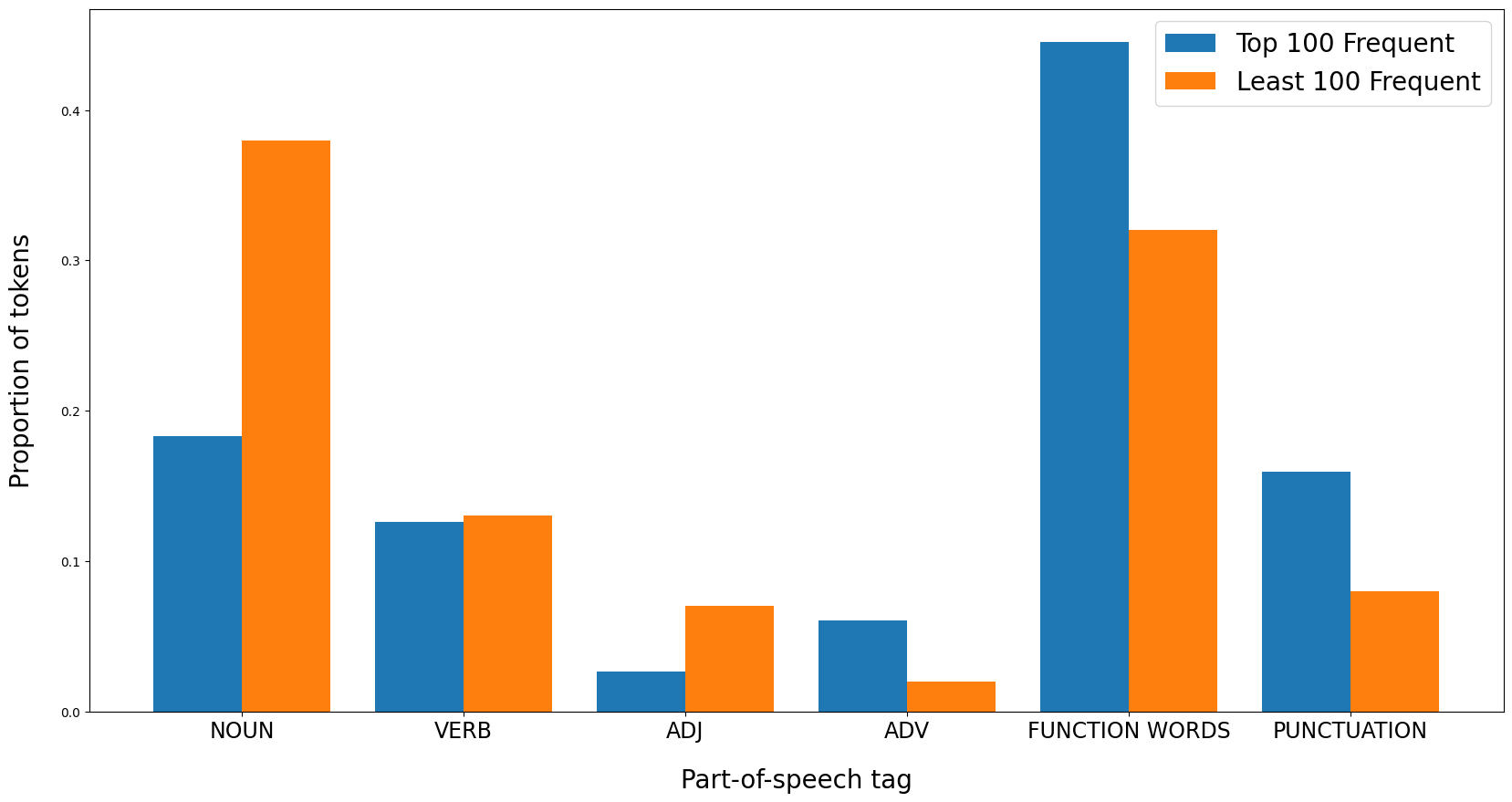}
    \caption{Distribution across POS tags of the top versus bottom 100 most frequent tokens.}
    \label{fig:top-100-pos-dist}
\end{figure}

\section{BLiMP Data Filtering}
\label{section:appendix-data-filtering}

We filter the BLiMP data to only focus on pairs of sentences where one set of tokens has been replaced by another set and ignore sentence pairs that only differ in the order of tokens. We also remove pairs where tokens have only been added to one sentence, rather than replaced. This filtering only removes 15\% of BLiMP pairs and 9 of the 67 subtasks from consideration. 

\section{NLU Performance of Open-Source Baselines}
\label{section:opensource-baselines-nlu-results}

\begin{table}[ht!]
\centering
\small
\setlength{\tabcolsep}{5pt} 
\begin{tabular}{l|c|cccc}
\toprule
\textbf{Model} & \textbf{BLiMP} & \textbf{COLA} & \textbf{SST-2} & \textbf{MNLI} & \textbf{QNLI} \\
\midrule
OPT & 63.2 & 64.6 & 81.9 & 57.6 & 61.5 \\
RoBERTa & 69.8 & 70.8 & 87.0 & 73.2 & 77.0 \\
T5 & 58.3 & 61.2 & 78.1 & 48.0 & 62.0 \\
\bottomrule
\end{tabular}
\caption{\label{tbl:opensource-baselines-nlu-results} BLiMP~($\uparrow$) score and accuracy ($\uparrow$) on sentence-level tasks (COLA, SST-2) and language inference tasks (MNLI, QNLI) for the three open-source transformer baselines.}
\end{table}

\end{document}